\documentclass[sigconf]{acmart}
\usepackage{multirow}
\usepackage{booktabs}
\usepackage{threeparttable}
\usepackage{algorithm}
\usepackage{enumitem}
\usepackage[noend]{algorithmic}
\usepackage{amsthm,amsmath,mathrsfs}
\usepackage{subfigure}
\usepackage{tikz}
\usepackage{xcolor}

\usepackage{subfigure}
\usepackage{color, colortbl}
\usepackage{makecell}
\usepackage{tikz}
\definecolor{Gray}{gray}{0.9}

\renewcommand{\algorithmiccomment}[1]{\bgroup\hfill{$\triangleright$\ }~#1\egroup}
\newcommand{\sstar}{$^{\boldsymbol{*}}$}
\AtBeginDocument{%
  \providecommand\BibTeX{{%
    \normalfont B\kern-0.5em{\scshape i\kern-0.25em b}\kern-0.8em\TeX}}}

\copyrightyear{2024}
\acmYear{2024}
\setcopyright{rightsretained}
\acmConference[CIKM '24]{Proceedings of the 33rd ACM International Conference on Information and Knowledge Management}{October 21--25, 2024}{Boise, ID, USA}
\acmBooktitle{Proceedings of the 33rd ACM International Conference on Information and Knowledge Management (CIKM '24), October 21--25, 2024, Boise, ID, USA}
\acmDOI{10.1145/3627673.3679554}
\acmISBN{979-8-4007-0436-9/24/10}

\makeatletter
\gdef\@copyrightpermission{
  \begin{minipage}{0.3\columnwidth}
   \href{https://creativecommons.org/licenses/by/4.0/}{\includegraphics[width=0.90\textwidth]{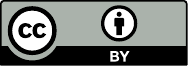}}
  \end{minipage}\hfill
  \begin{minipage}{0.7\columnwidth}
   \href{https://creativecommons.org/licenses/by/4.0/}{This work is licensed under a Creative Commons Attribution International 4.0 License.}
  \end{minipage}
  \vspace{5pt}
}
\makeatother

\begin{document}

\title{Prompt-Based Spatio-Temporal Graph Transfer Learning}
\author{Junfeng Hu}
\authornote{Work done when the author is a research intern at 
Institute for Infocomm Research, A*STAR, Singapore.}
\orcid{0000-0003-1409-1495}
\affiliation{%
  \institution{National University of Singapore}
  \country{Singapore}
}
\email{junfengh@u.nus.edu}

\author{Xu Liu}
\orcid{0000-0003-2708-0584}
\affiliation{%
  \institution{National University of Singapore}
  \country{Singapore}
}
\email{liuxu@comp.nus.edu.sg}

\author{Zhencheng Fan}
\orcid{0000-0002-8046-2541}
\affiliation{%
  \institution{University of Technology Sydney}
  \city{Sydney}
  \state{NSW}
  \country{Australia}
}
\email{zhencheng.fan@student.uts.edu.au}

\author{Yifang	Yin}
\orcid{0000-0002-6525-6133}
\affiliation{%
  \institution{Institute for Infocomm Research, A*STAR}
  \country{Singapore}
}
\email{yin_yifang@i2r.a-star.edu.sg}

\author{Shili Xiang}
\orcid{0000-0001-6598-2904}
\affiliation{%
  \institution{Institute for Infocomm Research, A*STAR}
  \country{Singapore}
}
\email{sxiang@i2r.a-star.edu.sg}

\author{Savitha	Ramasamy}
\orcid{0000-0003-1534-2989}
\affiliation{%
  \institution{Institute for Infocomm Research, A*STAR \& IPAL, CNRS@CREATE}
  \country{Singapore}
}
\email{ramasamysa@i2r.a-star.edu.sg}

\author{Roger Zimmermann}
\orcid{0000-0002-7410-2590}
\affiliation{%
  \institution{National University of Singapore}
  \country{Singapore}
}
\email{rogerz@comp.nus.edu.sg}
\renewcommand{\shortauthors}{Junfeng Hu et al.}

\begin{abstract}
Spatio-temporal graph neural networks have proven efficacy in capturing complex dependencies for urban computing tasks such as forecasting and kriging. Yet, their performance is constrained by the reliance on extensive data for training on a specific task, thereby limiting their adaptability to new urban domains with varied task demands. 
Although transfer learning has been proposed to remedy this problem by leveraging knowledge across domains, the cross-task generalization still remains under-explored in spatio-temporal graph transfer learning due to the lack of a unified framework.
To bridge the gap, we propose Spatio-Temporal Graph Prompting (STGP), a prompt-based framework capable of adapting to multi-diverse tasks in a data-scarce domain.
Specifically, we first unify different tasks into a single template and introduce a task-agnostic network architecture that aligns with this template. This approach enables capturing dependencies shared across tasks. 
Furthermore, we employ learnable prompts to achieve domain and task transfer in a two-stage prompting pipeline, facilitating the prompts to effectively capture domain knowledge and task-specific properties. Our extensive experiments demonstrate that STGP outperforms state-of-the-art baselines in three tasks—forecasting, kriging, and extrapolation—achieving an improvement of up to 10.7\%.
\end{abstract}





\maketitle

\section{Introduction}
The rise of urban computing has spurred the development of sensor technologies for monitoring urban conditions. The collected data, often represented as spatio-temporal graphs, has significantly changed our perception of understanding city environments, leading to various applications, such as forecasting~\cite{liu2023largest} and kriging~\cite{appleby2020kriging}. Recently, spatio-temporal graph neural networks (STGNNs) have stood as a dominant paradigm in urban computing, owing to their ability to capture complex spatio-temporal dependencies~\cite{wu2019graph,li2018diffusion,yu2018spatio,rao2022fogs,wu2021inductive,wu2021spatial}. The success of STGNNs primarily depends on their training within a singular task and dataset, which necessitates the availability of extensive data for effective optimization. However, this reliance on abundant data constrains their generalizability when a new urban domain with a multi-task requirement emerges~\cite{jin2023spatio}.

\begin{figure}
  \centering
  \includegraphics[width=0.96 \linewidth]{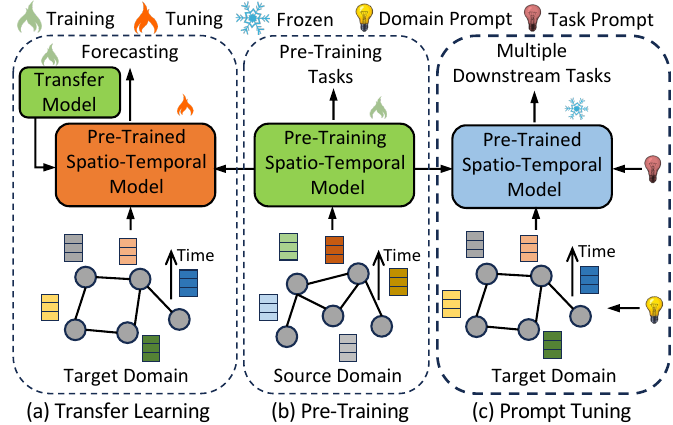}
  \caption{Comparison of existing transfer learning approaches with our prompt tuning. (a) The pre-trained model is adapted to a target domain, aided by a domain transfer model. (b) The spatio-temporal model is pre-trained on a data-rich source domain. (c) STGP leverages prompts to modify data, achieving domain and task transfer, respectively.}
  \vspace{-1em}
    \label{fig:intro}
\end{figure}

Transfer learning, which focuses on transferring knowledge across \emph{domains} and \emph{tasks}, emerges as a promising machine learning approach in scenarios where training data is limited~\cite{zhuang2020comprehensive}. It aims to leverage knowledge learned from a data-rich source domain to perform diverse downstream tasks in a data-scarce target domain. 
Nowadays, researchers have realized the data scarcity issue in spatio-temporal graphs and proposed a variety of solutions~\cite{wang2019cross,yao2019learning}. 
As shown in Fig.~\ref{fig:intro}(a)-(b), these models are first pre-trained on a source domain and then fine-tuned on a target domain. 
To achieve effective domain transfer, they are further supported by a transfer module.
For instance, CrossTReS~\cite{jin2022selective} captures domain shift by a weighting mechanism that filters out patterns exclusively in the source domain.
TPB~\cite{liu2023cross} stores shareable patterns across the domains by a memory bank. 
Despite the intriguing performance obtained by these systems, they are specifically designed for the task of forecasting by incorporating task-specific architectures. Consequently, they fall short in achieving task transfer due to their sensitivity to changes in task specifications ~\cite{brown2020language}. 
In this work, we aim to move towards general task-agnostic frameworks that are capable of performing multiple tasks in the target domain.


To tackle the problem of task transfer, an emerging technique known as prompt tuning has gained prominence~\cite{brown2020language}. This concept, originating from Natural Language Processing (NLP), first unifies different tasks into one task template with a single training objective. Then, it leverages pre-trained Transformers that are task-agnostic and employs prompts to instruct task-specific knowledge. Compared to fine-tuning, prompt tuning gains more efficiency by freezing the pre-trained model. Meanwhile, it still yields performance that is comparable to, or even surpasses the fine-tuning methods~\cite{liu2023pre}. Recently, the community has found that prompt tuning can also perform domain transfer by incorporating domain-specific prompts into the input data~\cite{ge2023domain}.
Inspired by this paradigm, we argue that prompting can similarly be introduced for transfer learning in spatio-temporal graphs, primarily for two key reasons.
First, prompting can suit different spatio-temporal tasks flexibly by leveraging a unified task template and task-agnostic architectures.
Second, it provides an innovative idea to achieve both domain and task transfer efficiently from a data perspective: keeping the model frozen while adapting the data space of the target domain.

However, applying prompt tuning to spatio-temporal graphs is non-trivial due to the following challenges. 
Firstly, to enable task transfer, the design of the model and the knowledge captured by it through pre-training should be task-agnostic. 
Taking language models~\cite{min2023recent} as an example, they unify the language tasks into a template of word prediction and utilize Transformers as the learning model. The template and the network architecture ensure its applicability and the learning of latent space shared by various tasks~\cite{qin2021exploring}. 
Nevertheless, \emph{how do we design a model and a task template for spatio-temporal graphs that preserves shareable dependencies among tasks} is under-explored. 
Secondly, existing prompting methods treat domain transfer and task transfer as distinct problems, introducing different solutions to address them by using either natural language or learnable prompts~\cite{ge2023domain,sun2023all}.
However, spatio-temporal graphs lack an effective way to incorporate language that can prompt domain and task knowledge.
Moreover, it is less apparent what mechanism the learnable prompts should take to guide both transfer objectives.
Thus, \emph{how to learn prompts that capture both domain knowledge and task-specific properties} is also a challenge waiting to be solved.

To address the aforementioned challenges, we propose a novel Spatio-Temporal Graph Prompting framework, termed STGP, with the goal of facilitating domain transfer while enabling capabilities for multiple downstream tasks.
First, we unify spatio-temporal graph tasks into a single task template, inspired by STEP~\cite{shao2022pre}, and introduce a task-agnostic network that aligns with this template. Moreover, we introduce a two-stage prompting approach. As depicted in Fig.~\ref{fig:intro}(c), the prompting method sequentially learns domain transfer and task transfer in its respective prompting stages.

\begin{figure}
  \centering
  \includegraphics[width=0.9 \linewidth]{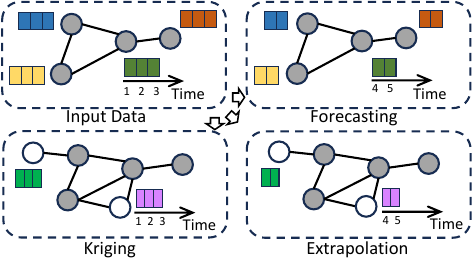}
  \vspace{-0.5em}
  \caption{Spatio-temporal graph downstream tasks.}
  \vspace{-1em}
    \label{fig:tasks}
\end{figure}

In particular, to address the first challenge, we unify spatio-temporal graph tasks into a data reconstruction template, where each task is represented as predictions of data at specific masked spatio-temporal positions.
To align with the template, we introduce a task-agnostic network, empowered by a Transformer-based encoder and a gated spatio-temporal decoder, that takes in input data to predict masked signals.
In this work, we adopt predictions of randomly masked data as the model's pre-training task to facilitate the learning of shareable spatio-temporal dependencies.
Then, as illustrated in Fig.~\ref{fig:tasks}, we apply the pre-trained model to two common downstream tasks—forecasting and kriging—and introduce a new task—extrapolation. The extrapolation task aims at predicting future signals for nodes lacking historical data, which holds significant practical importance. For example, it enables the use of limited sensors to forecast future traffic conditions across an entire city. 

Regarding the second challenge, we decouple the transfer of domain and task knowledge and propose a prompting pipeline consisting of two stages. The initial stage focuses on domain transfer, adapting the target domain's data space to the source domain used for model pre-training.
In practice, we freeze the pre-trained model, forward the domain-prompted data into the encoder, and train the prompts using the pre-training task. 
In the second stage, domain prompts are fixed and task prompts operate on the representations produced by the encoder. The task-prompted representations are fed into the decoder, enabling it to differentiate downstream tasks and reach optimal performance for each task. 
Our model is more efficient than current spatio-temporal transfer learning methods by leveraging domain prompts to transfer knowledge without the need for fine-tuning the pre-trained model. Meanwhile, by employing task prompting, our STGP shows versatility across multiple downstream tasks, yielding superior performance compared to other baselines. To summarize, our contributions lie in three aspects: 

\begin{itemize}[leftmargin=*]
    \item We introduce STGP, a novel framework for spatio-temporal graph transfer learning that leverages a unified template and a task-agnostic network to transfer knowledge across multiple tasks in the target domain, such as forecasting, kriging, and extrapolation. 
    \item We propose an innovative two-stage prompting strategy, empowering STGP to capture domain knowledge and task-specific properties for effective domain and task transfer.
    \item We conduct extensive experiments across four public datasets. The results and discussions highlight the superior performance of STGP and the efficacy of each component of the model.
\end{itemize}





\section{Related Work}
\subsection{Spatio-Temporal Graph Neural Networks} STGNNs have become one of the dominant paradigms for urban computing~\cite{jin2023spatio}, leading to a series of applications, such as forecasting and kriging. DCRNN~\cite{li2018diffusion} and STGCN~\cite{yu2018spatio} are two pioneer methods for forecasting, which employ GNNs to capture spatial relations and deploy sequential models to model temporal dependencies. Graph WaveNet~\cite{wu2019graph} proposes a matrix learning module to model hidden structural information, which becomes an influential idea for following methods~\cite{lan2022dstagnn,shao2022pre,rao2022fogs}. In terms of the task of kriging, a growing number of models have been recently developed to address it. For instance, KCN~\cite{appleby2020kriging} applies GNNs to capture dependencies between existing and unknown nodes, while IGNNK~\cite{wu2021inductive} further considers temporal relations. DualSTN~\cite{hu2023decoupling} learns joint correlations between time and space. SATCN~\cite{wu2021spatial} and INCREASE~\cite{zheng2023increase} propose to capture statistical graph information. 
However, STGNNs heavily rely on sufficient training data and are designed exclusively for a specific task. In this study, we are devoted to addressing these challenges through the proposed prompt-based framework.

\subsection{Spatio-Temporal Transfer learning} To improve the performance of STGNNs in scenarios with data scarcity, transfer learning methods are employed for spatio-temporal data. The model undergoes initial pre-training on a data-rich source domain and is then transferred to a target domain where data is scarce. Additionally, a transfer module is introduced to facilitate domain adaptation. RegionTrans~\cite{wang2019cross}, MetaST~\cite{pan2019urban}, and CrossTReS~\cite{jin2022selective} focus on grid data, learning region correlations between target and source cities. In the graph setting, ST-GFSL~\cite{lu2022spatio} introduces to generate network parameters specialized for the target domain. TPB~\cite{liu2023cross} learns temporal patterns shareable across the domains.  TransGTR~\cite{jin2023transferable} proposes to generate graph structures for the target domain that exhibit structural distributions akin to the source domain. Although these approaches introduce various transfer mechanisms to enhance model performance, most of them are optimized through a complex meta-learning framework such as Reptile~\cite{nichol2018first} or MAML~\cite{finn2017model} that require fine-tuning of the pre-trained model, which significantly diminishes their efficiency. Moreover, their exclusive focus on forecasting, using task-specific architectures, results in a lack of versatility for multiple downstream tasks.

\subsection{Prompt Tuning} Originating from Natural Language Processing, prompt tuning methods have gained widespread use for adapting pre-trained language models to a range of downstream tasks~\cite{liu2023pre}. These methods formulate the pre-training and downstream tasks into a single template of word prediction and employ learnable or natural language prompts to instruct task-related properties. In recent times, researchers have discovered that prompting shows potential in domain transfer by guiding the model with domain knowledge~\cite{ge2023domain}. Aside from the language tasks, prompting methods have also gained attention from various research areas such as Computer Vision~\cite{jia2022visual}, Graph Learning~\cite{sun2023all}, and Time Series Modeling~\cite{liu2023unitime,xue2023promptcast}. Despite their success, prompt-based methods remain under-explored in the context of spatio-temporal learning, with few proposed methods for graph~\cite{li2024flashst} and grid data~\cite{yuan2024unist} that focus on domain transfer only. Nevertheless, the transfer learning scenario demands simultaneous management of domain transfer and multi-task capabilities, thereby amplifying the complexity of the challenge.

\section{Preliminaries}
We outline the notations for spatio-temporal graphs, followed by formal definitions of transfer learning and the three specific downstream tasks we aim to address. 
\subsection{Data Notations}
\noindent\emph{Definition 1 (Graph)} Sensors can be organized as a graph represented by $\mathcal{G}=(\mathcal{V}, \mathcal{E}, A)$, where $\mathcal{V}$ is the set of nodes, $\mathcal{E}$ is the set of edges, and $A\in [0, 1]^{|\mathcal{V}|\times |\mathcal{V}|}$ is the normalized adjacency matrix describing the weights between nodes. 
\vspace{0.3em}

\noindent\emph{Definition 2 (Spatio-Temporal Data)} Signals gathered by sensors in the graph are represented as $X\in \mathbb{R}^{N\times T\times d_x}$, where $N$ indicates the number of nodes collecting data, $T$ represents a time window, and $d_x$ denotes the attribute count. We denote the data captured by node $i$ at time $t$ as $x_i^t \in \mathbb{R}^{d_x}$. 

\subsection{Problem Definitions}

\noindent\emph{Definition 3 (Spatio-Temporal Graph Transfer Learning)} Consider a model pre-trained on $P$ source domains with data over an substantial period $T_\mathcal{S}$ (e.g., one year), denoted as $\mathcal{G}^\mathcal{S}=\{\mathcal{G}_1^\mathcal{S},.. , \mathcal{G}_P^{\mathcal{S}}\}$. When a new target domain $\mathcal{G}^\mathcal{T}$ with $T_\mathcal{T} << T_\mathcal{S}$ (e.g., three days) emerges, the model is required to adapt to this new domain by utilizing knowledge acquired from the source domains and to perform various spatio-temporal graph learning tasks.
\vspace{0.3em}

\begin{figure*}[!h]
  \centering
  \includegraphics[width=0.97 \linewidth]{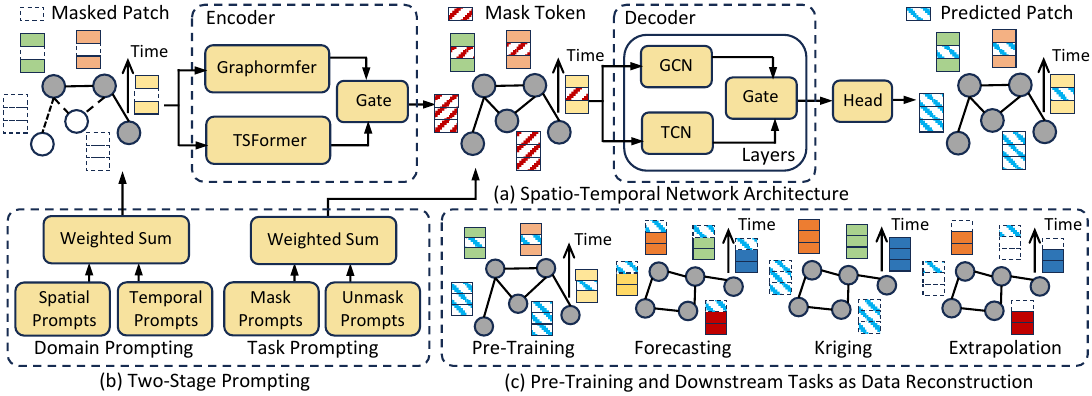}
  \caption{The overview of our Spatio-Temporal Graph Prompting framework. 
  (a) The network architecture is task-agnostic and aligns with the unified template. 
  (b) Domain and task prompts are learned sequentially through the two-stage prompting pipeline. (c) The tasks are consolidated within a data reconstruction template, where they are represented by masked locations.} 
  \vspace{-1em}
    \label{fig:framework}
\end{figure*}

\noindent\emph{Problem 1 (Forecasting)}
Given historical signals of $T$ time steps of $N$ nodes from a target domain, forecasting aims to transfer the pre-trained model $f(\cdot)$ to predict the data of subsequent $T^\prime$ steps:
\begin{equation}
    X^{1:T} \stackrel{f(\cdot)}{\longrightarrow} X^{T+1: T+T^\prime},
\end{equation}
where $X^{T+1: T+T^\prime}\in \mathbb{R}^{N\times T^\prime \times d_x}$ is the future readings. 
\vspace{0.3em}

\noindent\emph{Problem 2 (Kriging)}  
The goal of kriging is to adapt the model $f(\cdot)$ to predict data of $M$ unobserved locations given the $N$ observed nodes at the same time window $T$:
\begin{equation}
    X_{1:N} \stackrel{f(\cdot)}{\longrightarrow} X_{N+1:N+M},
\end{equation}
where $X_{N+1:N+M}\in \mathbb{R}^{M\times T \times d_x}$ is the data of unobserved locations.
\vspace{0.3em}

\noindent\emph{Problem 3 (Extrapolation)} Given historical signals from $N$ observed nodes, extrapolation involves using the model $f(\cdot)$ to predict $T^\prime$ future steps of $M$ unobserved locations:
\begin{equation}
    X_{1:N}^{1:T} \stackrel{f(\cdot)}{\longrightarrow} X_{N+1:N+M}^{T+1:T+T^\prime},
\end{equation}
where $X_{N+1:N+M}^{T+1:T+T^\prime} \in \mathbb{R}^{M\times T^\prime \times d_x}$ denotes the future data of unobserved locations.

\section{Methodology}

In this paper, we aim to introduce a spatio-temporal graph transfer learning framework, empowered by prompt tuning, that can be applied to multiple tasks. 
To achieve this goal, we first consolidate the pre-training and downstream tasks into a unified template as illustrated in Fig.~\ref{fig:framework}(c). Next, we design a network architecture that aligns with the template (Fig.~\ref{fig:framework}(a)) and propose a two-stage prompting pipeline for domain and task transfer (Fig.~\ref{fig:framework}(b)), respectively. 
In the following sections, we detail each component of STGP.

\subsection{Unification Spatio-Temporal Graph Tasks}
The effectiveness of applying a pre-trained model to diverse downstream tasks greatly relies on a unified task template, which ensures the learning of shareable latent space across tasks. 
%
However, it remains unclear how to unify these tasks for spatio-temporal graphs. As suggested by \cite{jin2023large}, different tasks place differing emphases on spatial and temporal dimensions. For instance, forecasting predominantly relies on the temporal evolution of historical signals, while kriging focuses more on the spatial relationships between observed and unobserved nodes. Consequently, existing models have to adopt task-specific training objectives and architectures to align with these properties~\cite{wu2021spatial,lu2022spatio}, which constrains their applicability on multiple downstream tasks. 

Inspired by STEP~\cite{shao2022pre} and GraphMAE~\cite{hou2022graphmae}, we reformulate the pre-training and downstream tasks as the reconstruction of masked data. 
As illustrated in Fig.~\ref{fig:framework}(c), this unified template involves predicting masked data that requires the extraction of useful spatio-temporal dependencies from unmasked signals.
In this way, these tasks can be defined when they are characterized by masks placed at specific positions.
For example, forecasting predicts masked signals at future time steps, while kriging aims to infer masked data that represents unobserved nodes. We adopt the predictions of randomly masked data as the pre-training task, which will be detailed below. 
Since the data masking relies solely on the task definition, without prioritizing spatial or temporal factors, our unified template ensures equal emphasis on dependencies across both dimensions, leading to consistent performance across all tasks.

\subsection{Pre-Training and Spatio-Temporal Networks}
Our pre-training approach involves predicting randomly masked data, propelling the model to identify shareable spatio-temporal dependencies for downstream tasks. 
To facilitate this, we introduce a spatio-temporal graph network featuring an autoencoder that comprises a Transformer-based encoder to capture complex dependencies within unmasked data and a gated decoder focusing on the predictions of masked data.

\subsubsection{Patching and Masking}
As individual node signals contain low information density, reconstructing each signal separately could lead to the learning of trivial solutions~\cite{shao2022pre}. This issue can be more acute in the transfer setting when a target domain involves limited data. Therefore, we employ data patching~\cite{liu2023cross} to aggregate adjacent data into patches. Concretely, given sequential data $X_i$ from node $i$, we divide it into non-overlapping patches denoted by $\hat{X}_i \in\mathbb{R}^{T_p\times L d_x}$, where $T_p$ is the number of patches and $L$ represents patch size. Then, we employ a linear projection to embed each patch into feature space, represented as $S_i\in\mathbb{R}^{T_p\times d_h}$. Note that we will use $T$ to denote $T_p$ for simplicity in the following sections.

Next, we pre-train the model on source domains by randomly masking the patches. Contrasting with the graph or time series models that utilize a noisy masking method~\cite{hou2022graphmae,liu2023unitime}, spatio-temporal random masking, which affects the two dimensions simultaneously, presents more complexity. As the Transformers only take in unmasked patches without padding~\cite{shao2022pre}, it is crucial to maintain a uniform patch number within each dimension to ensure a valid tensor structure, a requirement that noisy masking does not satisfy. Therefore, we divide the masking into spatial and temporal parts, where patches of a node or at a time step are all masked. Specifically, given a set of $N$ nodes and of $T$ steps, we randomly sample two sub-sets from them, denoted as $\mathcal{M}_s$ and $\mathcal{M}_t$. Then, a patch $s_i^t$ with its indexes in any of the sub-sets is marked as masked. The remaining unmasked patches, denoted by $S_{\bar{\mathcal{M}}}\in \mathbb{R}^{N_u\times T_u \times d_h}$ with ${S_{\bar{\mathcal{M}}}} = \{s_i^t| i\notin\mathcal{M}_s \text{ and } t\notin\mathcal{M}_t\}$, $N_u=N-|\mathcal{M}_s|$ and $T_u=T-|\mathcal{M}_t|$, are the inputs of the encoder. We set a total masking ratio of 75\%, following~\cite{shao2022pre,hou2023graphmae2}. The scarcity of unmasked patches in the model aids in better capturing general spatio-temporal correlations.

\subsubsection{Transformer-Based Encoder}
Transformers have become dominant in pre-training frameworks due to their attractive learning capabilities~\cite{khan2022transformers,kalyan2021ammus,wen2022transformers}. Additionally, masking further enhances their training efficiency on large-scale datasets by only feeding in unmasked patches. Nevertheless, employing a single Transformer to model both spatial and temporal dimensions still results in considerable time overhead, due to its complexity of $\mathcal{O}(N^2_u\times T^2_u)$. To overcome this issue, we model the two dimensions respectively by utilizing TSFormer~\cite{shao2022pre} to model temporal correlations and Graphormer~\cite{ying2021do} to capture spatial dependencies. 
Note that both models incorporate positional embeddings into their inputs.  
Following~\cite{jin2023transferable}, we enhance TSFormer by adding time of day (TOD) and day of week (DOW) embeddings, bolstering its capacity to capture periodic knowledge in the data-scarce target domain. In the bottom layer, representations learned from the two models are fused by a gating mechanism formulated as:
\begin{equation}
\begin{split}
H_{\bar{\mathcal{M}}} = Z \odot \operatorname{Enc_s}(S_{\bar{\mathcal{M}}}) + (1-Z) \odot \operatorname{Enc_t}(S_{\bar{\mathcal{M}}}),\\
Z = \sigma(\operatorname{Enc_s}(S_{\bar{\mathcal{M}}})W_1 + \operatorname{Enc_t}(S_{\bar{\mathcal{M}}})W_2 + b),
\end{split}
\label{eq:encoder}
\end{equation}
where $\operatorname{Enc_s}(\cdot)$, $\operatorname{Enc_t}(\cdot)$ are the two Transformers, $\sigma(\cdot)$, $\odot$ denote sigmoid and element-wise product, $W_1$, $W_2$, $b$ are learnable parameters. In this way, the time complexity of our encoder reduces to $\mathcal{O}(N^2_u + T^2_u)$, thereby achieving greater efficiency.
\subsubsection{Gated Spatio-Temporal Decoder}
The decoder first recovers full patches $H\in \mathbb{R}^{N\times T\times d_h}$ from $H_{\bar{\mathcal{M}}}$, where masked positions are replaced by a shared learnable mask token $h_\mathcal{M}$:
\begin{equation}
   h_i^t = \begin{cases} {h_{\bar{\mathcal{M}}}}_i^t & i\notin\mathcal{M}_s \text{ and } t\notin\mathcal{M}_t\ \\ 
   h_\mathcal{M} & i\in\mathcal{M}_s \text{ or } t\in\mathcal{M}_t\ \end{cases}.
\label{eq:unmask}
\end{equation}
Then, the decoder operates on $H$ to predict signals at masked positions. Here, the decoder architecture should be task-agnostic, ensuring its adaptability across all tasks.
%
Unfortunately, we cannot simply employ forecasting or kriging STGNNs as in other transfer learning frameworks~\cite{lu2022spatio,liu2023cross,jin2023transferable}. This limitation arises because most STGNNs prioritize the learning of either spatial or temporal dependencies, which can compromise their effectiveness on other tasks, as suggested by our experiments.
Thus, we propose a gated spatio-temporal graph decoder to circumvent the issue. Each decoder layer uses a graph convolutional network (GCN) and a temporal convolutional network (TCN) for parallel spatial and temporal modeling, followed by a gating mechanism for feature fusion. The process can be described by:
\begin{equation}
\begin{split}
H^l = Z \odot \operatorname{TCN}(H^{l-1}) + (1-Z) \odot \operatorname{GCN}(H^{l-1}, A),\\
Z = \sigma(\operatorname{TCN}(H^{l-1})W_1 + \operatorname{GCN}(H^{l-1}, A)W_2 + b),
\end{split}
\label{eq:decoder}
\end{equation}
where $l$ represents the layer index. In this way, the gating mechanism dynamically balances spatial and temporal patterns, facilitating the model on various tasks with different characteristics. Note that we still add the time embeddings at its first layer. Finally, we employ a linear prediction head to obtain patch predictions. We adopt mean absolute error (MAE) to optimize the model and calculate it solely on masked patches, in accordance with~\cite{he2022masked,shao2022pre}.

\subsection{Two-Stage Prompting}
In our transfer learning scenario, a pre-trained model needs to adapt to multiple downstream tasks in a target domain. 
Current prompting methods often introduce distinct solutions for domain transfer and task transfer~\cite{ge2023domain,sun2023all}. In contrast, we aim to achieve both goals within our single framework. Unfortunately, trivially optimizing a single bunch of prompts is inadequate for accomplishing the two goals. This is because domain transfer is designed to learn general domain knowledge, whereas task transfer focuses on capturing task-specific characteristics. Motivated by this observation, we introduce an innovative two-stage prompting pipeline that achieves these goals through a sequential, two-step approach.
\subsubsection{Prompt Bank}
We use a prompt bank as a basic prompting unit, where each bank contains a set of prompt vectors, defined as $P = (p_1, ..., p_{N_p})\in \mathbb{R}^{N_p\times d_h}$. The prompted vector for a given patch $h_i^t$ can be learned by a weighted sum:
\begin{align}
\begin{split}
\operatorname{PT}(h_i^t)=h_i^t + \Sigma_{j=1}^{N_p} \alpha_j p_j, \quad
\alpha_j=\begin{cases}
\sigma({h^t_i} p_j^\top) & \sigma({h^t_i} p_j^\top)>\varphi \\
0 & \sigma({h^t_i} p_j^\top) \leq\varphi
\end{cases},
\label{eq:prompting}
\end{split}
\end{align}
where $\varphi$ is a threshold and $\operatorname{PT(\cdot)}$ is the prompting function. Compared to methods such as single embedding prompting~\cite{liu2023graphprompt}, the prompt bank provides better abilities to transform patches in the latent space, which generally yields better performance~\cite{sun2023all}.
\subsubsection{Domain Prompting}
In the first stage, prompts are designed to capture the correlations between the domains and dynamically adapt the target domain data. To facilitate the capturing of domain knowledge and general spatio-temporal dependencies, we optimize the prompts by the same pre-training task with a large masking ratio and take the prompted patches as the encoder's inputs. 
We utilize different prompt banks for Graphormer and TSFormer, enabling them to transfer spatial and temporal domains, respectively. Note that positional embeddings are added to patches before prompting as prior knowledge. The process is defined as:
\begin{equation}
    S_s^* = \operatorname{PT_s}(S_{\bar{\mathcal{M}}} + PE_e), \quad S_t^* = \operatorname{PT_t}(S_{\bar{\mathcal{M}}} + PE_e),
\label{eq:dp}
\end{equation}
where $S_s^*$, $S_t^*$ are the prompted inputs of the two Transformers, $\operatorname{PT_s}(\cdot)$, $\operatorname{PT_t}(\cdot)$ are prompting functions, and $PE_e$ are positional embeddings of the encoder. In this stage, only the prompts are trainable while the other modules are frozen without fine-tuning, which increases efficiency over other transfer learning methods. 

\subsubsection{Task Prompting} The task prompts are learned respectively for each downstream task to prompt out task-specific properties.
We incorporate the prompts into the hidden representations learned by the encoder. Then, the decoder takes in the task-prompted representations to predict target signals at masked positions based on the learned target domain dependencies. 
We utilized separate prompt banks for mask tokens and unmasked patches, thereby facilitating the capturing of relationships between them:
\begin{equation}
    H^*_\mathcal{M} = \operatorname{PT_m}(H_\mathcal{M} + PE_d), \quad H^*_{\bar{\mathcal{M}}} = \operatorname{PT_u}(H_{\bar{\mathcal{M}}} + PE_d),
\label{eq:tk}
\end{equation}
where $\operatorname{PT_m}(\cdot)$ and $\operatorname{PT_u}(\cdot)$ are promptings for mask tokens and unmasked patches, $PE_d$ is the embeddings of the decoder. Note that in this stage, we further freeze the domain prompts. As downstream tasks feature different masked position patterns, task prompting can benefit the model by identifying these distinctions, thereby achieving optimal performance for each task.

\subsection{STGP Learning Procedure} We present the overall learning algorithm of STGP in Algo.~\ref{algo:framework}. Given the encoder $f_{\theta_{e}}$, the decoder $f_{\theta_{d}}$, and the head $f_\phi$ parameterized by $\theta_{e}$, $\theta_{d}$, and $\phi$, we first train them through predictions of randomly masked patches. Then, keeping them frozen, the domain prompts $\pi_{dm}$ are optimized using the pre-training task for domain transfer. Finally, further with $\pi_{dm}$ held fixed, task prompts $\pi_{tk}$ are trained for a task. The head can optionally be tuned at this stage.

\begin{algorithm}
\caption{STGP Pre-Training and Prompting}
\begin{algorithmic}[1]
\REQUIRE Source and target domain data $\mathcal{G}^\mathcal{S}$, $\mathcal{G}^\mathcal{T}$. Encoder $f_{\theta_e}$, decoder $f_{\theta_d}$, head $f_\phi$. Domain and task prompts $\pi_{dm}$ and $\pi_{tk}$.
\ENSURE Optimal parameters $\theta_{e}$, $\theta_{d}$, $\phi$, $\pi_{dm}$ and $\pi_{tk}$.
\STATE // Pre-training $f_{\theta_e}$, $f_{\theta_d}$ and $f_\phi$ with $\mathcal{G}^\mathcal{S}$
\WHILE{not converged}
\STATE $S_{\bar{\mathcal{M}}}, S_{\mathcal{M}} \longleftarrow \text { PatchingMasking }(\mathcal{G}^\mathcal{S})$
\STATE $\hat{S}_{\mathcal{M}} = f_{\phi, \theta_{e}, \theta_{d}}(S_{\bar{\mathcal{M}}})$ \COMMENT{Data reconstruction, Eq.~\ref{eq:encoder} -~\ref{eq:decoder}}
\STATE Calculate loss $\operatorname{MAE}(\hat{S}_{\mathcal{M}}, S_{\mathcal{M}})$ and optimize $\theta_e$, $\theta_d$, and $\phi$.
\ENDWHILE
\STATE // Domain prompting using $\mathcal{G}^\mathcal{T}$
\STATE $f_{\theta_{e}}, f_{\theta_{d}}, f_\phi \longleftarrow $ Freeze $(\theta_{e}, \theta_{d}, \phi)$
\WHILE{not converged}
\STATE $S_{\bar{\mathcal{M}}}, S_{\mathcal{M}} \longleftarrow \text { PatchingMasking }(\mathcal{G}^\mathcal{T})$
\STATE $S_{\bar{\mathcal{M}}}^* \longleftarrow$ DomainPrompting $(S_{\bar{\mathcal{M}}}, \pi_{dm})$ \COMMENT{Eq.~\ref{eq:dp}}
\STATE $\hat{S}_{\mathcal{M}} = f_{\phi, \theta_{e}, \theta_{d}}(S_{\bar{\mathcal{M}}}^*)$ \COMMENT{Eq.~\ref{eq:encoder} -~\ref{eq:decoder}}
\STATE Calculate loss and optimize $\pi_{dm}$. 
\ENDWHILE
\STATE // Task prompting 
\STATE $f_{\theta_{e}}, f_{\theta_{d}}, f_\phi, \pi_{dm} \longleftarrow $ Freeze $(\theta_{e}, \theta_{d}, \phi, \pi_{dm})$
\WHILE{not converged}
\STATE $S_{\bar{\mathcal{M}}}, S_{\mathcal{M}} \longleftarrow \text { PatchingMasking }(\mathcal{G}^\mathcal{T})$ \COMMENT{Downstream task masking}
\STATE $S_{\bar{\mathcal{M}}}^* \longleftarrow$ DomainPrompting $(S_{\bar{\mathcal{M}}}, \pi_{dm})$ \COMMENT{Eq.~\ref{eq:dp}}
\STATE $H_{\bar{\mathcal{M}}}=f_{\theta_{e}}(S^*_{\bar{\mathcal{M}}})$ \COMMENT{Eq.~\ref{eq:encoder}}
\STATE $H \longleftarrow$ ReocverFullPatches $(H_{\bar{\mathcal{M}}})$ \COMMENT{Eq.~\ref{eq:unmask}}
\STATE $H^* \longleftarrow$ TaskPrompting $(H, \pi_{tk})$ \COMMENT{Eq.~\ref{eq:tk}}
\STATE $\hat{S}_{\mathcal{M}} = f_{\phi, \theta_{d}}(H^*)$ \COMMENT{Eq.~\ref{eq:decoder}}
\STATE Calculate loss and optimize $\pi_{tk}$ \COMMENT{$\phi$ can be tuned optionally}
\ENDWHILE
\RETURN Trained models $f_{\theta_{e}}, f_{\theta_{d}}, f_{\phi}$ and prompts $\pi_{dm}, \pi_{tk}$
\end{algorithmic}
\label{algo:framework}
\end{algorithm}



\section{Experiment}

\begin{table*}[!h]
\centering
\caption{Forecasting performance of transfer learning on the four datasets. Each target city only has limited data, whereas the remaining three datasets are regarded as the source domain. The best and second-best are indicated in bold and underlined. Marker * indicates statistical significance, denoting that the p-value from the Student’s T-test is less than 0.01.}
\vspace{-1em}
  \scalebox{0.85}{
  \centering
  \begin{threeparttable}
  \begin{tabular}[width=1.\linewidth]{lccccccccc cccccccc}
    \toprule
      \multirow{3}*{\textbf{Model}} & \textbf{Target City} & \multicolumn{8}{c}{\textbf{METR-LA}}  & \multicolumn{8}{c}{\textbf{PEMS-BAY}}\\
      \cmidrule(r){3-6} \cmidrule(r){7-10} \cmidrule(r){11-14} \cmidrule(r){15-18}
      & \textbf{Metrics} & \multicolumn{4}{c}{\textbf{MAE ($\downarrow$)}} & \multicolumn{4}{c}{\textbf{RMSE ($\downarrow$)}} & \multicolumn{4}{c}{\textbf{MAE ($\downarrow$)}} & \multicolumn{4}{c}{\textbf{RMSE ($\downarrow$)}}\\
      & \textbf{Horizon} & 15 m & 30 m & 60 m & avg. & 15 m & 30 m & 60 m & avg. & 15 m & 30 m & 60 m & avg. & 15 m & 30 m & 60 m & avg.\\
      \midrule
      HA & \multirow{4}*{Target Only} & 3.62 & 4.23 & 5.13 & 4.26 & 7.33 & 8.56 & 10.17 & 8.54 & 2.42 & 2.89 & 3.70 & 2.93 & 5.46 & 6.52 & 8.25 & 6.59 \\
      ARIMA &  & 3.22 & 3.70 & 5.00 & 3.65 & 6.28 & 7.80 & 9.80 & 7.78 & 2.28 & 2.49 & 3.66 & 2.47 & 4.52 & 5.35 & 7.45 & 5.36 \\
      DCRNN &  & 3.03 & 3.73 & 4.81 & 3.77 & 5.78 & 7.34 & 9.34 & 7.27 & 1.81 & 2.36 & 3.24 & 2.38 & 3.39 & 4.75 & 6.37 & 4.70 \\
      GWN &  & 3.08 & 3.70 & 4.73 & 3.69 & 5.79 & 7.30 & 9.10 & 7.31 & 1.94 & 2.40 & 3.08 & 2.40 & 3.60 & 4.67 & 6.19 & 4.67 \\
      \midrule
      DCRNN & \multirow{5}*{Reptile} & 3.01 & 3.65 & 4.67 & 3.69 & 5.62 & 7.16 & 8.96 & 7.03 & 1.83 & 2.43 & 3.36 & 2.44 & 3.36 & 4.71 & 6.59 & 4.68 \\
      GWN &  & 3.11 & 3.75 & 4.73 & 3.76 & 5.87 & 7.31 & 9.10 & 7.21 & 1.99 & 2.45 & 3.14 & 2.55 & 3.55 & 4.64 & 6.23 & 4.65\\
      DSTAGNN &  & 3.30 & 4.10 & 4.95 & 4.12 & 5.90 & 7.73 & 9.56 & 7.74  & 1.85 & 2.51 & 3.59 & 2.48 & 3.41 & 4.79 & 6.66 & 4.80 \\
      FOGS & & 3.26 & 4.11 & 4.88 & 4.14 & 5.95 & 7.50 & 9.47 & 7.51 & 1.89 & 2.38 & 3.37 & 2.29 & 3.49 & 4.54 & 6.01 & 3.52 \\
      STEP &  & 3.20 & 3.99 & 4.76 & 4.00 & 5.70 & 7.14 & 8.53 & 7.12 & 1.72 & 2.30 & 3.08 & 2.31 & 3.21 & 4.38 & 5.83 & 4.40 \\
      \midrule
      AdaRNN & \multirow{5}*{Transfer} & 3.05 & 3.68 & 4.51 & 3.66 & 5.66 & 7.15 & 8.60 & 7.10 & 1.79 & 2.33 & 3.04 & 2.38 & 3.38 & 4.60 & 5.98 & 4.58 \\
      ST-GFSL & & \underline{3.00} & 3.79 & 4.58 & 3.80 & 5.72 & 7.21 & 8.67 & 7.18 & 1.77 & 2.20 & 2.95 & 2.24 & 3.27 & 4.50 & 5.92 & 4.45 \\
      DSATNet & & 3.03 & 3.66 & 4.51 & 3.63 & 5.70 & 7.15 & 8.78 & 7.08 & 1.64 & 2.16 & 2.88 & 2.18 & 3.26 & 4.36 & 5.89 & 4.30\\
      TPB &  & 3.07 & 3.80 & 4.66 & 3.83 & 5.69 & \underline{7.03} & 8.52 & 6.95 & \underline{1.62} & \textbf{2.12} & 2.83 & 2.15 & 3.24 & \underline{4.33} & 5.76 & \underline{4.23}\\
      TransGTR &  & 3.01 & \underline{3.64} & \underline{4.44} & \underline{3.62} & \underline{5.60} & 7.12 & \underline{8.49} & \underline{6.89} & \textbf{1.60} & \underline{2.13} & \underline{2.79} & \underline{2.14} & \textbf{3.04} & 4.35 & \underline{5.68} & 4.26 \\
      \midrule
      STGP & \multirow{2}*{Transfer} & \textbf{2.97}\sstar & \textbf{3.54}\sstar & \textbf{4.23}\sstar & \textbf{3.52}\sstar & \textbf{5.48}\sstar & \textbf{6.77}\sstar & \textbf{8.19}\sstar & \textbf{6.67}\sstar & 1.74$^*$ & \underline{2.13} & \textbf{2.70}\sstar & \textbf{2.12} & \underline{3.21}$^*$ & \textbf{4.18}\sstar & \textbf{5.46}\sstar & \textbf{4.14}\sstar\\
      Std. &  & .003 & .010 & .008 & .009 & .029 & .011 & .019 & .011 & .017 & .014 & .012 & .011 & .013 & .015 & .014 & .017 \\
      \midrule
      \midrule
      \multirow{3}*{\textbf{Model}} & \textbf{Target City} & \multicolumn{8}{c}{\textbf{Chengdu}}  & \multicolumn{8}{c}{\textbf{Shenzhen}}\\
      \cmidrule(r){3-6} \cmidrule(r){7-10} \cmidrule(r){11-14} \cmidrule(r){15-18}
      & \textbf{Metrics} & \multicolumn{4}{c}{\textbf{MAE ($\downarrow$)}} & \multicolumn{4}{c}{\textbf{RMSE ($\downarrow$)}} & \multicolumn{4}{c}{\textbf{MAE ($\downarrow$)}} & \multicolumn{4}{c}{\textbf{RMSE ($\downarrow$)}}\\
      & \textbf{Horizon} & 15 m & 30 m & 60 m & avg. & 15 m & 30 m & 60 m & avg. & 15 m & 30 m & 60 m & avg. & 15 m & 30 m & 60 m & avg.\\
      \midrule
      HA & \multirow{4}*{Target Only} & 2.69 & 3.13 & 3.65 & 3.01 & 3.69 & 4.57 & 5.27 & 4.40 & 2.17 & 2.66 & 3.04 & 2.54 & 3.33 & 4.05 & 4.63 & 3.89 \\
      ARIMA &  & 2.97 & 3.28 & 4.32 & 3.19 & 3.78 & 4.64 & 5.51 & 4.58 & 2.35 & 2.99 & 3.60 & 2.90 & 4.32 & 4.73 & 5.58 & 4.62 \\
      DCRNN &  & 2.27 & 3.05 & 3.60 & 2.82 & 3.26 & 4.38 & 5.07 & 4.03 & 2.05 & 2.61 & 3.10 & 2.49 & 3.17 & 3.98 & 4.60 & 3.79\\
      GWN &  & 2.25 & 2.96 & 3.42 & 2.74 & 3.20 & 4.26 & 4.90 & 3.94 & 1.86 & 2.46 & 2.84 & 2.27 & 2.76 & 3.69 & 4.28 & 3.41\\
      \midrule
      DCRNN & \multirow{5}*{Reptile} & 2.31 & 3.16 & 3.96 & 2.97 & 3.33 & 4.55 & 5.42 & 4.28 & 1.94 & 2.57 & 3.07 & 2.38 & 2.84 & 3.81 & 4.52 & 3.53\\
      GWN &  & 2.19 & 2.81 & 3.21 & 2.62 & 3.12 & 4.08 & 4.65 & 3.79 & 1.88 & 2.46 & 2.82 & 2.27 & 2.77 & 3.68 & 4.26 & 3.40\\
      DSTAGNN &  & 2.36 & 3.01 & 3.40 & 2.85 & 3.21 & 4.20 & 4.97 & 3.99 & 1.98 & 2.43 & 2.95  & 2.42 & 2.90 & 3.69 & 4.27 & 3.38 \\
      FOGS &  & 2.23 & 2.80 & 3.31 & 2.61 & 3.18 & 4.30 & 4.77 & 4.05 & 1.96 & 2.36 & 2.80 & 2.14 & 2.88 & 3.61 & 4.35 & 3.33 \\
      STEP &  & 2.21 & 2.70 & 3.20 & 2.49 & 3.05 & 3.99 & 4.29 & 3.76 & 1.95 & 2.40 & 2.55 & 2.17 & 2.73 & 3.44 & 3.80 & 3.22\\
      \midrule
      AdaRNN & \multirow{4}*{Transfer} & 2.18 & 2.91 & 3.40 & 2.73 & 3.09 & 3.97 & 4.82 & 3.68 & 1.92 & 2.48 & 2.88 & 2.28 & 2.85 & 3.63 & 4.26 & 3.38 \\
      ST-GFSL &  & 2.10 & 2.80 & 3.35 & 2.59 & 3.02 & 3.88 & 4.60 & 3.57 & 1.90 & 2.36 & 2.71 & 2.19 & 2.70 & 3.53 & 4.19 & 3.26 \\
      DSATNet & & 2.06 & 2.70 & 3.03 & 2.46 & 3.02 & 4.01 & 4.53 & 3.66 & 1.86 & 2.34 & 2.64 & 2.14 & 2.73 & 3.51 & 4.00 & 3.22 \\
      TPB &  & 2.08 & \underline{2.63} & 3.02 & 2.45 & 2.98 & 3.84 & 4.34 & 3.52 & \underline{1.85} & 2.32 & 2.61 & \underline{2.13} & 2.70 & \underline{3.45} & 3.91 & 3.20 \\
      TransGTR &  & \underline{2.05} & 2.65 & \underline{2.80} & \underline{2.41} & \underline{2.95} & \underline{3.82}\underline & \underline{4.26} & \underline{3.49} & 1.89 & \underline{2.30} & \underline{2.47} & 2.15 & \underline{2.69} & 3.49 & \underline{3.79} & \underline{3.19}\\
      \midrule
      STGP & \multirow{2}*{Transfer} & \textbf{1.98}\sstar & \textbf{2.54}\sstar & \textbf{2.74}\sstar & \textbf{2.30}\sstar & \textbf{2.85}\sstar & \textbf{3.72}\sstar & \textbf{4.02}\sstar & \textbf{3.37}\sstar & \textbf{1.82}\sstar & \textbf{2.27}\sstar & \textbf{2.42}\sstar & \textbf{2.08}\sstar & \textbf{2.66}\sstar & \textbf{3.39}\sstar & \textbf{3.69}\sstar & \textbf{3.11}\sstar\\
      Std. & & .012 & .010 & .010 & .014 & .011 & .010 & .016 & .013 & .008 & .012 & .011 & .013 & .017 & .013 & .010 & .012\\
  \bottomrule
\end{tabular}
\end{threeparttable}
}
\label{tab:forecasting}
\end{table*}

\subsection{Experimental Settings}

\noindent \textbf{Datasets.}
We evaluate our STGP using four public traffic flow datasets: METR-LA, PEMS-BAY~\cite{li2018diffusion}, Chengdu, and Shenzhen~\cite{dididataset}. 

\noindent \textbf{Baselines.}
We consider 19 baselines belonging to three categories. 
\begin{itemize}[leftmargin=*]
\item \textbf{Statistical Methods:} HA, MEAN, KNN, ARIMA~\cite{williams2003modeling} calculate the statistical properties of input data to predict node signals. 
\item \textbf{Typical Neural Networks:} DCRNN~\cite{li2018diffusion}, GWN~\cite{wu2019graph}, FOGS~\cite{rao2022fogs}, DSTAGNN~\cite{lan2022dstagnn}, STEP~\cite{shao2022pre}, KCN~\cite{appleby2020kriging}, IGNNK~\cite{wu2021inductive}, SATCN~\cite{wu2021spatial}, DualSTN~\cite{hu2023decoupling}, and INCREASE~\cite{zheng2023increase} are models for forecasting or kriging. To apply them in the transfer learning scenario, we optimize them using the Reptile~\cite{nichol2018first} meta-learning framework.
\item \textbf{Transfer Learning Models:} AdaRNN~\cite{du2021adarnn}, ST-GFSL~\cite{lu2022spatio}, DSATNet~\cite{tang2022domain}, TPB~\cite{liu2023cross}, TransGTR~\cite{jin2023transferable} are state-of-the-art time series or spatio-temporal graph methods for transfer learning.
\end{itemize}

\noindent \textbf{Evaluation Settings and Implementation.}
We adopt the same setting as previous works~\cite{lu2022spatio,liu2023cross}, where the datasets are divided into source, target, and test datasets. The source datasets comprise three datasets for pre-training. We sequentially split them into training and validation segments with a ratio of $8:2$, where the model is selected on the validation part. 
Then only the first three-day data in the remaining dataset is regarded as the target dataset for domain and task prompting. The subsequent three-day data is utilized for model selection and the rest of the segment is adopted to compute metrics.
We adopt mean absolute error (MAE) and root mean squared error (RMSE) as the metrics. 

Hyperparameters are the same in all settings for STGP. Each patch, representing one-hour data, has a length of $L=12$, with $T=25$ patches in total. 
Both Graphormer and TSFormer contain 4 layers with a hidden size of 128. The decoder contains 6 spatio-temporal gated layers with a hidden size of 32. We also add skip and residual connections into the decoder. The prediction head is a two-layer linear projection with a size of 128 and 256. All prompt banks involve 25 vectors with a size of 128. 
For extrapolation, we fine-tune the prediction
head when optimizing the task prompts, as we empirically find it
can yield better results.
We run each experiment 5 times to report results. Our model is implemented by PyTorch and trained on one NVIDIA A100 GPU with 32GB memory. The source code is available at \url{https://github.com/hjf1997/STGP}.

\begin{table*}[h]
\centering
\caption{Kriging performance of transfer learning on the datasets. Trdu./Indu.  denote transductive/inductive settings.}
\vspace{-1em}
  \scalebox{0.85}{
  \centering
  \begin{threeparttable}
  \begin{tabular}[width=1.\linewidth]{lccccc cccc cccc cccc}
    \toprule
      \multirow{3}*{\textbf{Model}} & \textbf{Target City} & \multicolumn{4}{c}{\textbf{METR-LA}}  & \multicolumn{4}{c}{\textbf{PEMS-BAY}} & \multicolumn{4}{c}{\textbf{Chengdu}}  & \multicolumn{4}{c}{\textbf{Shenzhen}}\\
      \cmidrule(r){3-4} \cmidrule(r){5-6} \cmidrule(r){7-8} \cmidrule(r){9-10} \cmidrule(r){11-12} \cmidrule(r){13-14} \cmidrule(r){15-16} \cmidrule(r){17-18}
      & \textbf{Metrics} & \multicolumn{2}{c}{\textbf{MAE ($\downarrow$)}} & \multicolumn{2}{c}{\textbf{RMSE ($\downarrow$)}} & \multicolumn{2}{c}{\textbf{MAE ($\downarrow$)}} & \multicolumn{2}{c}{\textbf{RMSE ($\downarrow$)}} & \multicolumn{2}{c}{\textbf{MAE ($\downarrow$)}} & \multicolumn{2}{c}{\textbf{RMSE ($\downarrow$)}} & \multicolumn{2}{c}{\textbf{MAE ($\downarrow$)}} & \multicolumn{2}{c}{\textbf{RMSE ($\downarrow$)}}\\
      & \textbf{Setting} & Trdu. & Indu. & Trdu. & Indu. & Trdu. & Indu. & Trdu. & Indu. & Trdu. & Indu. & Trdu. & Indu. & Trdu.  & Indu. & Trdu. & Indu.\\
      \midrule
      MEAN & \multirow{2}*{Target Only} & \multicolumn{2}{c}{10.28} & \multicolumn{2}{c}{14.86} & \multicolumn{2}{c}{5.49} & \multicolumn{2}{c}{9.28} & \multicolumn{2}{c}{7.42} & \multicolumn{2}{c}{9.78} & \multicolumn{2}{c}{8.81} & \multicolumn{2}{c}{11.08} \\
      KNN &  & \multicolumn{2}{c}{9.11} & \multicolumn{2}{c}{12.88} & \multicolumn{2}{c}{4.20} & \multicolumn{2}{c}{7.55} & \multicolumn{2}{c}{7.01} & \multicolumn{2}{c}{9.40} & \multicolumn{2}{c}{8.89} & \multicolumn{2}{c}{11.17} \\
      \midrule
      KCN & \multirow{5}*{Reptile} & 5.69 & 7.76 & 8.72 & 11.78 & 3.89 & 4.76 & 6.91 & 8.20 & 6.64 & 7.03 & 8.99  & 9.31 & 5.70  & 8.28 & 8.08 & 10.98\\
      IGNNK & & 5.67 & 7.60 & 9.33 & 11.68 & 3.94 & 4.30 & 7.16 & 7.22 & 6.71 & 7.56 & 8.90 & 9.94 & 5.41 & 8.20 & \underline{7.58} & 10.90\\
      SATCN & & \underline{4.81} & 7.51 &\underline{7.85} & 11.59 & 3.48 & 3.85 & 6.51 & 6.84 & 6.61 & 7.14 & 9.03 & 9.42  & 5.53 & \underline{8.15} & 7.89 & 10.87\\
      DualSTN & & 5.62 & \underline{7.12} & 9.60 & \underline{11.46} & 3.54 & 3.74 & 6.82 & \underline{6.75} & \underline{5.95} & \underline{6.42} & \underline{8.07} & \underline{8.71} & 5.37 & 8.18 & 7.66 & 10.91\\
      INCREASE & & 5.03 & 7.38 & 8.24 & 11.54 & \underline{3.46} & \underline{3.72} & \underline{6.41} & 6.78 & 6.14 & 6.63 & 8.53 & 8.80 & \underline{5.34} & 8.16 & 7.61 & \underline{10.85}\\
      \midrule
      
      STGP & \multirow{2}*{Transfer} & \textbf{4.53}\sstar & \textbf{6.98}\sstar & \textbf{7.75}\sstar & \textbf{11.39} & \textbf{3.39}\sstar & \textbf{3.66}\sstar & \textbf{6.28}\sstar & \textbf{6.68}\sstar & \textbf{5.47}\sstar & \textbf{6.33}\sstar & \textbf{7.70}\sstar & \textbf{8.64} & \textbf{5.25}\sstar & \textbf{8.08}\sstar & \textbf{7.50} & \textbf{10.79}\\
      Std. & & .022 & .062 & .037 & .081 & .013 & .033 & .018 & .040 & .033 & .055 & .014 & .053 & .019 & .083  & .049 & .088\\
  \bottomrule
\end{tabular}
\end{threeparttable}
}
\label{tab:kriging}
\end{table*}

\begin{table*}
\vspace{-1em}
\centering
\caption{Extrapolation performance of transfer learning on the datasets. The results are averaged over all 12 future horizons.}
\vspace{-1em}
  \scalebox{0.85}{
  \centering
  \begin{threeparttable}
  \begin{tabular}[width=1.\linewidth]{lccccc cccc cccc cccc}
    \toprule
      \multirow{3}*{\textbf{Model}} & \textbf{Target City} & \multicolumn{4}{c}{\textbf{METR-LA}}  & \multicolumn{4}{c}{\textbf{PEMS-BAY}} & \multicolumn{4}{c}{\textbf{Chengdu}}  & \multicolumn{4}{c}{\textbf{Shenzhen}}\\
      \cmidrule(r){3-4} \cmidrule(r){5-6} \cmidrule(r){7-8} \cmidrule(r){9-10} \cmidrule(r){11-12} \cmidrule(r){13-14} \cmidrule(r){15-16} \cmidrule(r){17-18}
      & \textbf{Metrics} & \multicolumn{2}{c}{\textbf{MAE ($\downarrow$)}} & \multicolumn{2}{c}{\textbf{RMSE ($\downarrow$)}} & \multicolumn{2}{c}{\textbf{MAE ($\downarrow$)}} & \multicolumn{2}{c}{\textbf{RMSE ($\downarrow$)}} & \multicolumn{2}{c}{\textbf{MAE ($\downarrow$)}} & \multicolumn{2}{c}{\textbf{RMSE ($\downarrow$)}} & \multicolumn{2}{c}{\textbf{MAE ($\downarrow$)}} & \multicolumn{2}{c}{\textbf{RMSE ($\downarrow$)}}\\
      & \textbf{Setting} & Trdu. & Indu. & Trdu. & Indu. & Trdu. & Indu. & Trdu.  & Indu. & Trdu. & Indu. & Trdu. & Indu. & Trdu.  & Indu. & Trdu. & Indu.\\
      \midrule
      MEAN & Target Only & \multicolumn{2}{c}{10.64} & \multicolumn{2}{c}{15.27} & \multicolumn{2}{c}{5.56} & \multicolumn{2}{c}{9.31} & \multicolumn{2}{c}{7.53} & \multicolumn{2}{c}{9.74} & \multicolumn{2}{c}{8.91} & \multicolumn{2}{c}{11.63}\\
      \midrule
      GWN & \multirow{4}*{Reptile} & 5.41 & 9.26 & 9.69 & 15.10 & 4.32 & 5.01 & 8.09 & 9.86 & 5.83 & 7.21 & 8.25 & 9.61 & 4.02 & 8.47 & 5.74 & 11.40\\
      DCRNN & & \underline{5.10} & 9.41 & \underline{9.19} & 15.57 & \underline{3.64} & 5.12 & \underline{6.97} & 10.14 & 6.23 & 7.10 & 8.62 & 9.58 & \underline{3.85} & 8.64 & \underline{5.31} & 11.54\\
      IGNNK &  & 6.67 & 9.14 & 10.33 & {14.63} & 3.90 & 5.05 & 7.13 & 9.90 & 6.52 & 7.43 & 8.81 & 9.71 & 4.41 & 8.52 & 6.47 & 11.38 \\
      SATCN & & 7.29 & 9.13 & 12.41 & 14.88 & 4.65 & 5.13 & 9.04 & 10.13 & 7.09 & 7.20 & 9.51 & 9.66  & 4.87 & 8.70 & 7.15 & 11.58\\
      \midrule
      ST-GFSL &  \multirow{3}*{Transfer} & 7.73 & 9.31 & 12.31 & 14.99 & 4.27 & 4.59 & 7.61 & 7.92 & 6.51 & 6.79 & 8.74 & 9.41  & 4.75 & 8.55 & 6.99 & 11.45\\
      TPB & & 7.97 & \underline{9.01} & 12.50 & 14.42 & 4.40 & \underline{4.53} & 7.79 & \underline{7.83} & \underline{6.12} & 6.67  & \underline{8.37} & 9.27 & 4.66 & \underline{8.38} & 6.96 & \underline{11.23} \\
      TransGTR & & 7.33 & 8.87 & 11.22 & \underline{14.12} & 4.55 & 4.66 & 7.95 & 8.10 & 6.42 & \underline{6.62} & 8.65 & \underline{9.10} & 4.58 & 8.45 & 6.81 & 11.36\\
      \midrule
      STGP & \multirow{2}*{Transfer} & \textbf{5.04} & \textbf{8.51}\sstar & \textbf{9.03}\sstar & \textbf{13.49} & \textbf{3.41}\sstar & \textbf{4.13}\sstar & \textbf{6.47}\sstar & \textbf{7.52}\sstar & \textbf{5.46}\sstar & \textbf{6.51}\sstar & \textbf{7.67} & \textbf{8.91}\sstar & \textbf{3.61}\sstar & \textbf{8.22}\sstar & \textbf{5.04}\sstar & \textbf{11.07} \\
      Std. & & .083 & .094 & .084 & .159 & .010 & .045 & .021 & .088 & .019 & .086 & .024 & .091 & .057 & .051 & .074 & .090\\
  \bottomrule
\end{tabular}
\end{threeparttable}
}

\label{tab:extrapolation}
\end{table*}

\subsection{Performance Evaluation}

\textbf{Forecasting.} We conduct forecasting using the similar setting of~\cite{liu2023cross}, where 1-day historical data (i.e., 24 patches) is utilized to predict subsequent signals of 1 hour (i.e., 1 patch). To assess Reptile's impact on spatio-temporal transfer learning, we present the results of two baselines directly trained within the target domain. Table~\ref{tab:forecasting} reports the metrics, from which we have the following observations. (1) STGP consistently attains superior results on multiple datasets, reducing the metrics of the second-best by up to 5.6\% on MAE and 4.5\% on RMSE. This performance demonstrates the effectiveness of STGP. (2) TransGTR performs better on some metrics. The probable reason is that it utilizes knowledge distillation to transfer long-term temporal relations from the source domain, which is absent in our model. (3) Simply equipping a model with Reptile only leads to marginal improvement. This highlights the need for transfer learning systems in spatio-temporal graphs amid data scarcity.

\noindent \textbf{Kriging.} We conduct experiments on both transductive and inductive settings for kriging. 
In the transductive setting, the model has access to the structure of unobserved nodes during training, thereby presenting a greater challenge.
Conversely, the inductive setting secludes these nodes throughout the training and utilizes them solely for evaluation.
We set the ratio of observed and unobserved nodes to $2:1$ and report the results in Table~\ref{tab:kriging}. It shows that (1) Our STGP consistently obtains the best results, reducing MAE and RMSE by up to 7.0\% and 8.0\%. This improvement underscores its superiority. (2) Several neural networks struggle to outperform statistical models on the datasets. This highlights the challenge of kriging when models predict nodes without any signals amidst limited training data. 

\noindent \textbf{Extrapolation.} We conduct evaluations following the same experimental settings.
As a new task, we compare STGP against compatible baselines. As Table~\ref{tab:extrapolation} shows, our analysis reveals that (1) STGP consistently surpasses the baselines across metrics by a significant margin. In particular, STGP achieves a reduction in MAE and RMSE by up to 8.8\% and 10.7\%, marking the highest performance among the tasks. (2) STGP yields higher standard divisions on extrapolation and kriging. As they require predicting node signals solely based on graph structures, the restricted spatial information poses challenges for the model in avoiding local minima during the optimization.

\subsection{Ablation Study}
\begin{figure}[!b]
  \centering
  \includegraphics[width=0.98 \linewidth]{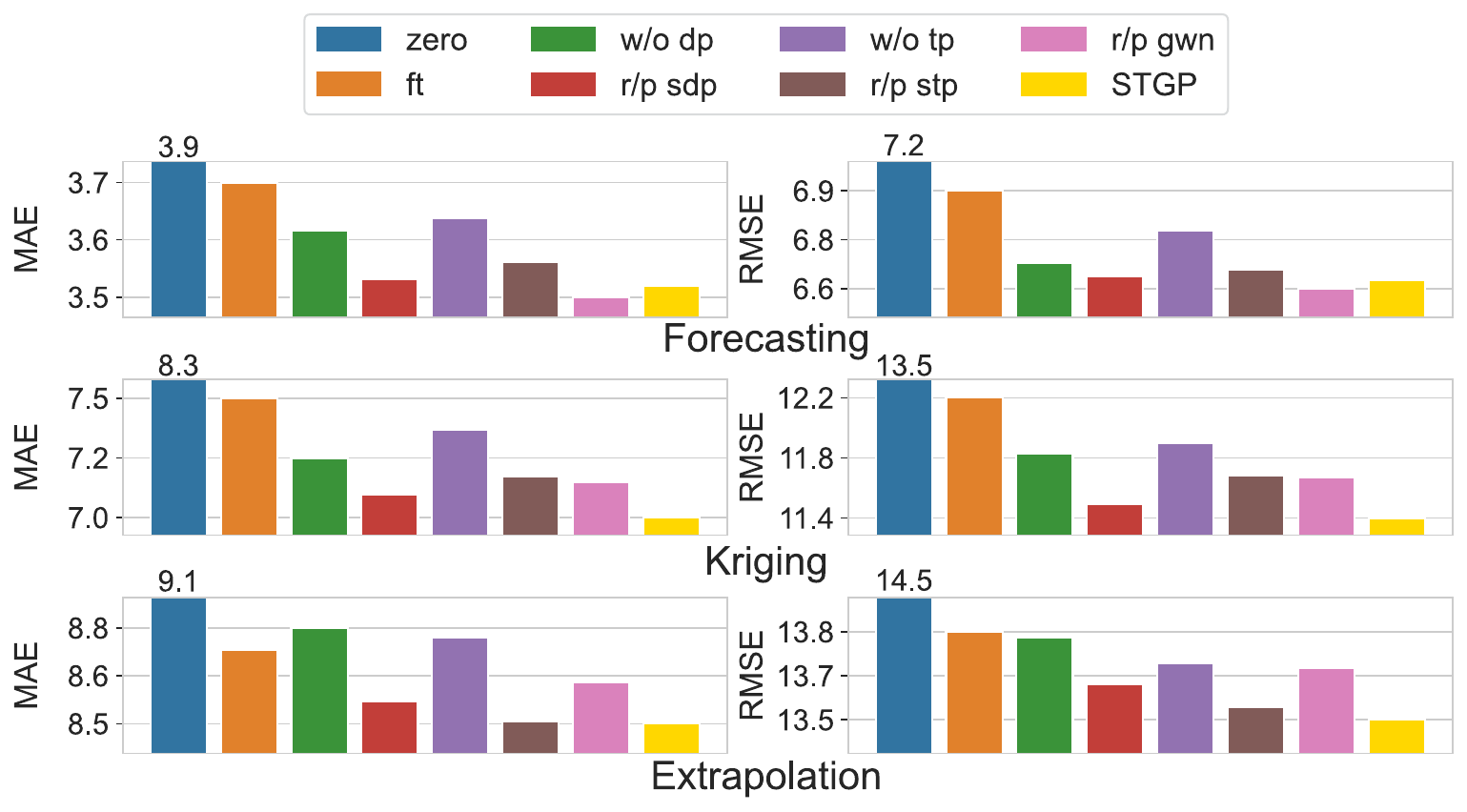}
  \vspace{-1em}
  \caption{Model analysis with METR-LA as the target domain.}
    \label{fig:ablation}
\end{figure}

We conduct ablation studies on the three tasks by using METR-LA as the target domain. For kriging and extrapolation, we adopt the inductive setting. The results are reported in Fig.~\ref{fig:ablation}.

\noindent \textbf{Prompt Tuning.}
We remove all prompts, evaluating the pre-trained model on the target domain directly (i.e., zero-shot setting) or fine-tuning it, denoted as \textbf{zero} and \textbf{ft}. We observed that both settings yield worse performance. This suggests that domain shift impedes a pre-trained model from directly adapting to a target domain. Meanwhile, the effectiveness of fine-tuning is adversely affected due to limited target data and discrepancies between pre-training and downstream task objectives.
On the contrary, STGP unifies the tasks, using prompts for effective domain and task transfer. 

\noindent \textbf{Domain Prompting.}
We remove temporal and spatial prompt banks (\textbf{w/o dp}) or replace them with a single bank (\textbf{r/p sdp}). Our results show that separate banks achieve the best performance. The explanation is that the two dimensions involve domain shifts with different characteristics, which should be modeled individually. 

\noindent \textbf{Task Prompting.}
We remove task prompts (\textbf{w/o tp}) or replace them with a single bank (\textbf{r/p stp}). It shows that STGP requires both banks for optimal results. This is because masked patches are filled by a mask token, which is different from patches with extracted dependencies from the encoder. The two prompt banks benefit the decoder in predicting masked patches based on captured relations.

\noindent \textbf{Gated Decoder.} We analyze the efficacy of our gated decoder by substituting it with a task-specific model, Graph WaveNet (\textbf{r/p gwn}). This variant shows comparable forecasting results but experiences a significant deterioration in other tasks. As the design of these forecasting models prioritizes capturing of temporal dependencies~\cite{wu2019graph,yu2018spatio,lan2022dstagnn}, the absence of temporal relations in the rest tasks—owing to the replacement of all target nodes with mask tokens—impedes their effectiveness. The problem can also appear when applying kriging methods~\cite{hu2023decoupling,wu2021spatial} to forecasting. Conversely, our decoder places equal importance on both spatial and temporal dimensions and dynamically fuses these dependencies via a gating mechanism, which renders it adaptable to a variety of tasks.

\subsection{Efficiency Analysis} To assess the efficiency of STGP's prompting, we compute its complexities. Suppose a bank contains $N_p$ prompts with dimension $d_h$, the total parameters for the two prompting stages are $\mathcal{O}(4N_pd_h)$. For time complexity, assume the model has $L$ layers, the encoder's complexity is approximated as $\mathcal{O}(LTN^2d_h+LNT^2d_h)$. As the decoder is a combination of GCN and TCN, the complexity is $\mathcal{O}(LN^2d_h+LTkd_h^2)$, where $k$ is the kernel size. The prompting by Eq.~\ref{eq:prompting} incurs an extra cost of $\mathcal{O}(3N_pTNd_h)$, which is minimal given $N_P \ll LT \ll LN$. We observe that, in terms of dataset size, the parameter cost is invariant while the time cost scales linearly. 

To empirically compare our model with transfer learning baselines, we present the number of tunable parameters and floating point operations (FLOPs) for forecasting. We set the batch size to 1 and report FLOPs of one iteration. 
As shown in Table~\ref{tab:efficiency}, the results reveal that TransGTR, requiring training of TSFormer on the target domain, incurs the most cost. 
ST-GFSL, employing a generator for parameter prediction, also shows high computational demand. TPB, which tunes WaveNet, has the least demand among baselines.  
In contrast, STGP not only exceeds the performance of these baselines but also significantly reduces the need for parameters and FLOPs, by factors of approximately 100 and 25, respectively.

\begin{table}
\caption{Tunable parameters and FLOPs comparison. STGP-dp and STGP-tp represent domain and task prompting.}
\vspace{-1.em}
  \scalebox{0.87}{
  \centering
  \begin{tabular}[width=0.76\linewidth]{lccccc}
    \toprule
       \multirow{2}*{Methods} & \multirow{2}*{\makecell{Params \\ $\times 10^3$}} & \multicolumn{4}{c}{FLOPs $\times 10^9$} \\
       \cmidrule(r){3-6}
       & & METR-LA & PEMS-BAY & Chengdu & Shenzhen\\
       \midrule
       ST-GFSL & 878 & 10.82 & 19.55 & 38.49 & 50.36\\
       TPB & 567 & 10.52 & 19.13 & 37.92 & 49.75\\
       TransGTR & 1,137 & 16.48 & 28.79 & 54.99 & 71.51\\ 
       \midrule
       STGP-dp & 3 & 0.16 & 0.27 & 0.43 & 0.51\\
       STGP-tp & 3 & 0.33 & 0.52 & 0.84 & 1.00\\
  \bottomrule
\end{tabular}
}
\label{tab:efficiency}
\end{table}

\subsection{Hyperparameter Study} 
\begin{figure}
  \centering
  \includegraphics[width=1 \linewidth]{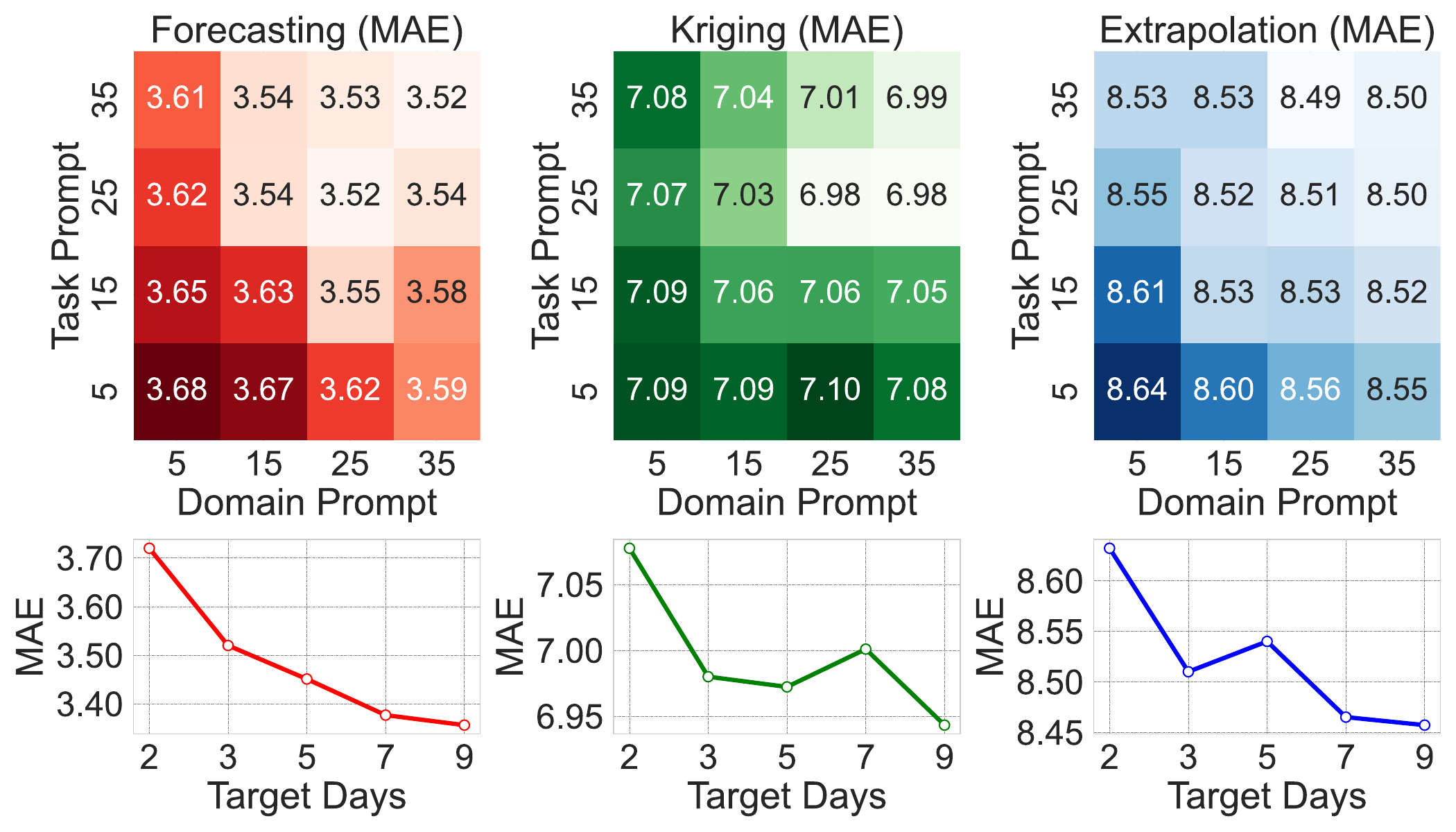}
  \vspace{-2.em}
  \caption{Effect of prompt numbers and the number of target days for training on the METR-LA dataset.}
    \label{fig:num_prompts}
\end{figure}

We first evaluate the number of domain and task prompts, using METR-LA as the target domain. The first row of Fig.~\ref{fig:num_prompts} displays the MAE results, with lighter colors denoting superior performance. The results reveal that optimal performance requires a substantial number of both prompts. A large number of one prompt type coupled with a small number of the other type hinders the model's performance. This finding suggests distinct roles for the two prompts: domain prompts address domain transfer, while task prompts focus on task-specific properties. Next, we discuss the effect of available data on the target domain. As the second row shows, the MAE decreases with the increase of training data. This suggests that data scarcity is clearly a bottleneck for model performance, necessitating the introduction of effective transfer learning methods like STGP.


\subsection{Domain Transfer Analysis}
\begin{figure}
  \centering
  \includegraphics[width=0.92 \linewidth]{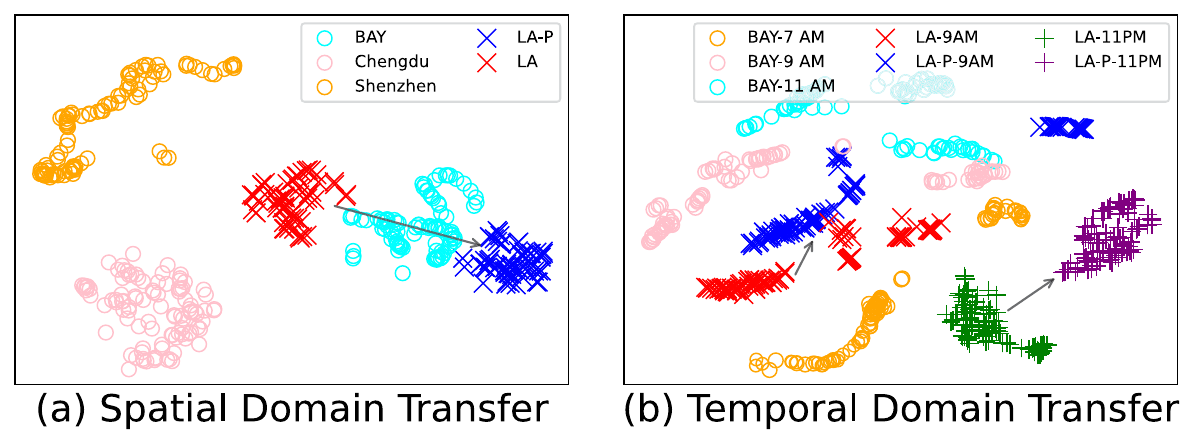}
  \vspace{-1.em}
  \caption{Visualization of domain transfer. $\circ$ represents source domain; $\times$, $+$ denote target domain. The arrows indicate the representation shift of a target domain.}
  \vspace{-1.em}
    \label{fig:vis_la}
\end{figure}

We innovatively utilize prompt tuning to achieve domain transfer for spatio-temporal graphs. To analyze its efficacy, we use METR-LA as the target domain and visualize hidden representations of patches. For spatial domain transfer, we randomly select 100 nodes from each of the datasets and average representations of a node at all time steps. As shown in Fig.~\ref{fig:vis_la} (a), after prompting (\textbf{LA-P}), the METR-LA points (\textbf{LA}) shift away from Chengdu and Shenzhen while staying close to PEMS-BAY (\textbf{BAY}). This might be explained by the proximity between Los Angeles and the Bay area. For temporal transfer, we average nodes at the same time and visualize points at different times. Fig.~\ref{fig:vis_la} (b) shows that METR-LA points at 9 AM move towards PEMS-BAY at 9 AM, while the 11 PM points show no clear trend. As the model is pre-trained on a source domain, these findings suggest that domain prompting can effectively transform target data to align with the latent space of the source domain.

\section{Conclusion}
In this paper, we present STGP, a novel prompt-based approach for spatio-temporal graph transfer learning. We develop a unified template alongside a task-agnostic network architecture suitable for multiple tasks. We then propose a two-stage prompting pipeline tailored for domain and task transfer. Comprehensive experiments and discussions demonstrate that STGP effectively captures domain knowledge and task-specific properties, consistently achieving state-of-the-art performance across diverse datasets and tasks.
Regarding future work, we find that the downstream tasks unified by STGP fall under predictive learning, where known data is utilized to predict unknown data in a regression setting. Although received less attention, spatio-temporal graphs also encompass other applications, such as classification. Exploring the applicability of our unified template and network architecture to these tasks presents an intriguing direction for future research.

\begin{acks}
This research is supported by Singapore Ministry of Education Academic Research Fund Tier 2 under MOE's official grant number T2EP20221-0023. It is also part of the programme DesCartes and is supported by the National Research Foundation, Prime Minister’s Office, Singapore under its Campus for Research Excellence and Technological Enterprise (CREATE) programme.

\end{acks}

\appendix

\bibliographystyle{ACM-Reference-Format}
\balance
\bibliography{sample-base}

\end{document}